# Monotone Tree-Based GAMI Models by Adapting XGBoost[1]

Linwei Hu, Soroush Aramideh, Jie Chen, Vijayan N. Nair

Corporate Model Risk, Wells Fargo, USA

**Abstract**

Recent papers have used machine learning architecture to fit low-order functional ANOVA models with main effects and second-order interactions. These GAMI (GAM + Interaction) models are directly interpretable as the functional main effects and interactions can be easily plotted and visualized. Unfortunately, it is not easy to incorporate the monotonicity requirement into the existing GAMI models based on boosted trees, such as EBM (Lou et al. 2013) and GAMI-Lin-T (Hu et al. 2022). This paper considers models of the form $f(x) = \sum_{j,k} f_{j,k}(x_j, x_k)$ and develops monotone tree-based GAMI models, called monotone GAMI-Tree, by adapting the XGBoost algorithm. It is straightforward to fit a monotone model to $f(x)$ using the options in XGBoost. However, the fitted model is still a black box. We take a different approach: i) use a filtering technique to determine the important interactions, ii) fit a monotone XGBoost algorithm with the selected interactions, and finally iii) parse and purify the results to get a monotone GAMI model. Simulated datasets are used to demonstrate the behaviors of mono-GAMI-Tree and EBM, both of which use piecewise constant fits. Note that the monotonicity requirement is for the full model. Under certain situations, the main effects will also be monotone. But, as seen in the examples, the interactions will not be monotone.

**Keywords:** generalized additive models with interactions; inherently interpretable models; functional ANOVA; machine learning; monotonicity

## 1. Introduction

Early work in machine learning (ML) with tabular data focused on complex algorithms to get the best predictive performance. The need for interpretability, particularly in highly regulated industries, is now leading researchers to revisit the need for unnecessary model complexity and instead consider trade-offs between predictive performance and interpretability. Recent work uses ML architecture (boosted trees and neural networks with special structure) to fit models based on lower-order functional ANOVA (fANOVA) framework with functional main effects and second-order interactions:

$$f(x) = \sum_{j} f_j(x_j) + \sum_{j \neq k} f_{jk}(x_j, x_k). \qquad Eq\ (1)$$

This class of models was referred to as GA2M in Lou et al. (2013) but as GAMI (Generalized Additive Models + Interactions) by others. We will use the GAMI terminology. Explainable boosting machine (EBM), in Lou et al. (2013), uses gradient boosting with piecewise constant trees to fit GAMI models. GAMI-Linear-Tree (GAMI-Lin-T), proposed in Hu et al. (2022), applies boosting with specialized linear model-based trees, and it has several other key differences with EBM. GAMI-Net, developed by Yang et al. (2021), uses (specialized) neural network structures and the associated optimization techniques to fit the

---

GAMI models. They are all multi-stage algorithms that involve fitting the functional main effects and interactions in stages and use techniques to filter the interactions and select the most important ones. We refer readers to these papers for more details. In particular, Hu et al. (2023) provides comparisons of the three algorithms in terms of predictive performance and model interpretability. Interestingly, in empirical analyses, these GAMI models do not seem to lose much predictive performance while being directly interpretable. Hence, they are becoming popular in regulated industries such as banking.

An additional key requirement for models is that they satisfy the monotonicity property for suitable variables. For example, in credit scoring applications, the fitted probability of default should be non-increasing as credit age increases (see, for example, Nair et al. (2022) for more context). Unfortunately, it is not easy to incorporate the monotonicity requirement in EBM and GAMI-Lin-T due to the multi-stage algorithm design.

There are several approaches in the literature to incorporate monotonicity of ML models. One class of methods imposes soft monotonicity constraints, where a penalty term is added to the loss function to penalize monotonicity violations. See, for example, Liu et al. (2020) for certified monotonic neural networks and Kokel et al. (2020) for knowledge-intensive gradient boosting. The disadvantage of soft monotone techniques is that they do not guarantee full monotonicity. A second class of methods builds in hard monotonicity constraints. These include monotone XGBoost, Isotonic Classification Trees (Bonakdarpour et al. 2018), Monotone Rule RF (Bartley et al. 2019), and Tensorflow Lattice. Yang et al. (2021), the authors of the GAMI-Net method, have since adapted it to enforce monotonicity using Tensorflow Lattice.

This paper develops monotone GAMI models based on boosted trees. Instead of trying to retrofit EBM and GAMI-Lin-T, which seems very challenging, the paper adapts the monotone XGBoost algorithm in the literature. There are several steps in turning the black-box monotone XGBoost into a monotone GAMI model, and they are described in Section 2.

The rest of the paper is organized as follows. We describe this algorithm in detail in Section 2, followed by simulation studies in Section 3. Finally, we make some concluding remarks in Section 4.

## 2. Monotone GAMI-Tree

As noted earlier, we adapt the monotone XGBoost algorithm to get our Monotone GAMI-Tree model. The original XGBoost is computationally efficient, and it comes with many features (options) such as regularization. More importantly for our purpose, it also has options to include monotonicity and two-way interaction constraints (specifying which interactions to fit). As mentioned in Bartley et al. (2019), the monotonicity implementation in XGBoost is ideal "because the trees are shallow (depth ≈ 3), and any bias introduced by one tree can be corrected by subsequent trees". Therefore, it is a natural choice to use XGBoost model with two-way interaction and monotonicity constraints instead of trying to monotonize EBM (note the documentation on the XGBoost website[2] has some misleading information regarding interaction constraint. See the Appendix A for details).

However, the fitted monotone XGBoost model is still a black box. To achieve the transparency of GAMI, we must decompose the fitted Xgboost model as a fANOVA model in Eq (1). Here are the steps involved in fitting monotone GAMI-Tree. Each of these steps are further elaborated below.

---

[2] https://xgboost.readthedocs.io/en/stable/tutorials/feature_interaction_constraint.html

1) Interaction Filtering:
    a) Fit an initial GAM model to estimate the main effects;
    b) Compute the residuals after fitting the GAM model (see Appendix B);
    c) Use the interaction filtering algorithm in EBM (Lou et al. 2013) or GAMI-Lin-Tree (Hu et al. 2022) to select the top K pairwise interactions.
    d) The parameter K can be treated as a hyper-parameter and selected by hyper-parameter tuning.
2) Fit a monotone XGBoost model with main effects and the specified K second-order interactions;
3) Parse the fitted results to separate the main effects and second-order interactions; and
4) Purify the main effects and interactions to ensure they satisfy the hierarchical orthogonal property.

Details:

Step 1: The GAM model can be fitted using any one of the available algorithms. It is not necessary to keep it monotone as this is just an initial fit. However, if needed, one can use monotone XGBoost with depth of one which will then fit only main effects. The residuals after fitting GAM model are used to filter interactions. For the continuous response case, residual is simply defined as $y - \hat{f}(x)$. For binary response, we use the pseudo-residual definition discussed in Appendix B. The interaction filtering algorithm in EBM, aka FAST, uses a simple 4-quadrant model to fit interaction effects, whereas in GAMI-Lin-Tree, it uses a model-based tree. The benefit of using model-based tree is that it is more flexible, hence can capture interaction effects better. However, FAST works reasonably well in general. The goal in Step 1) is to reduce the number of interactions being fitted and parsed in Steps 2) and 3). Alternatively, one could skip Step 1) and replace Step 2) by fitting a monotone XGBoost with depth two which will constrain the fitted model to at most second-order interactions. One can then follow Steps 3) and 4). However, this will involve parsing potentially all $\binom{p}{2}$ interactions in Step 3 (even if one were to prune the unimportant ones later) and can be time consuming. In addition, there is a chance that an XGBoost algorithm with depth two does not capture all interactions (as some of the splits in the second level may be spend on capturing non-linearity). In our experience, this is unlikely, but it needs to be investigated.

Step 2: As noted earlier, there are options in XGBoost to restrict the pairwise interactions that are to be fit and the variables that should be monotonic.

Step 3: To parse the Xgboost model, aggregate each main-effect and interaction term. Specifically, for each tree leaf, we check if it is modeling main-effect or interaction and update the corresponding main-effect or interaction term. The details are given in Algorithm 1.

| **Algorithm 1: Parsing** |
| --- |
| Initialize $f_j(x_j) = 0$, $f_{jk}(x_j, x_k) = 0$ |
| For each tree $\mathcal{T}$ in Xgboost: |
|   For each leaf node $\mathcal{N}$ in $\mathcal{T}$: |
|     Get the set of split variables for node $\mathcal{N}$: $x_j$ or $(x_j, x_k)$ |
|     Get the split rule $I(x_j \in \mathcal{N})$ or $I(x_j, x_k \in \mathcal{N})$, and node value $a$ |
| Update $f_j(x_j) = f_j(x_j) + \lambda a I(x_j \in \mathcal{N})$, or $f_{j,k}(x_{j,k}) = f_{j,k}(x_{j,k}) + \lambda a I(x_j, x_k \in \mathcal{N})$, $\lambda$ is the learning rate. |

<u>Step 4</u>: Once parsing is done, we get an initial decomposition of $f(x) = \sum_j f_j(x_j) + \sum_{j,k} f_{jk}(x_j, x_k)$. However, the interaction term $f_{jk}(x_j, x_k)$ from this stage may not be orthogonal to the main effect, which will give misleading interpretation. This can happen because in XGBoost, a tree can split on $x_j, x_k$ purely due to their main effects instead of interaction. To resolve this issue, we purify the interaction terms by removing any part of the main effects. This will ensure the interaction and main-effect terms are hierarchically orthogonal (Hooker 2007). Here we use a linear B-spline model, but other GAM models can be used as well. The details are given in Algorithm 2.

| **Algorithm 2: Purification** |
|---|
| *Input: train set, $f_j(x_j), f_{jk}(x_j, x_k)$* |
| For each term $f_{jk}(x_j, x_k)$: |
|     Get the prediction on train set: $\tilde{y}_i = f_{jk}(x_{ij}, x_{ik})$ |
|     Fit a main-effect model $\tilde{y} \sim g_j(x_j) + g_k(x_k)$ |
|     Update $f_j(x_j) = f_j(x_j) + g_j(x_j), f_k(x_k) = f_k(x_k) + g_k(x_k)$, $f_{j,k}(x_{j,k}) = f_{j,k}(x_{j,k}) - g_j(x_j) - g_k(x_k)$ |

The complete description of monotone GAMI-Tree algorithm is given in Algorithm 3.

| **Algorithm 3: Mono-GAMI-Tree** |
|---|
| *Input: train, valid, k* |
| Use Step 1) above to identify the top $k$ interaction pairs; |
| Tune a monotone Xgboost model with the identified interactions; |
| Parse the fitted Xgboost model to separate main effects and interactions; |
| Purify the main effects and interactions to be orthogonal; and |
| Compute measures of importance for the term and visualize their effects (see Section 3 for examples). |

**Remarks:**
- One can replace the interaction filtering part in Step 1) with any other algorithm in the literature.
- Another key difference with EBM is that this method does not use two-stage training of main effects and interactions. Hu et al. (2022) showed that the two-stage training method in EBM may not converge to the best model unless the main effects and interactions are uncorrelated.

## 3. Simulation Results

We compare monotone GAMI-Tree algorithm with EBM through two simulation models. The first one is a first-order model

$$f(x) = 0.5x_1 + x_2 I(x_2 > 0) + x_3 I(x_3 < 0) + 0.5 \frac{\exp(6x_4) - 1}{1 + \exp(6x_4)},$$

with all the $x'_j s$ are independent Unif$(-1,1)$. This is a GAM model with four variables, each having different shapes, and all of them are monotone increasing. For continuous case, we take $y = f(x) + \epsilon$, where $\epsilon$ is $N(0, \sigma^2 = 4)$, and for binary case, $logit\ P(Y = 1|x) = f(x)$.

We simulated 15,000 observations, divided them into training (50%), validation (25%), and testing sets (25%). The training set was used to train the model and validation set to determine the best hyper-parameters. For the hyper-parameters, we assumed the number of interactions K is known as this is not a primary aspect of the simulation study. We tuned number of trees, max_depth, learning rate, etc., for monotone GAMI-Tree and number of bins, learning rate, min_samples_leaf, etc., for EBM.

*Table 1. Performance for EBM and monotone GAMI-Tree for First-Order Model*

|  | **Continuous** | | **Binary** | |
| --- | --- | --- | --- | --- |
| **Algorithm** | **Train (rmse)** | **Test (rmse)** | **Train (AUC)** | **Test (AUC)** |
| mono-GAMI-Tree | 1.991 | 1.984 | 0.688 | 0.670 |
| EBM | 1.993 | 1.985 | 0.691 | 0.668 |

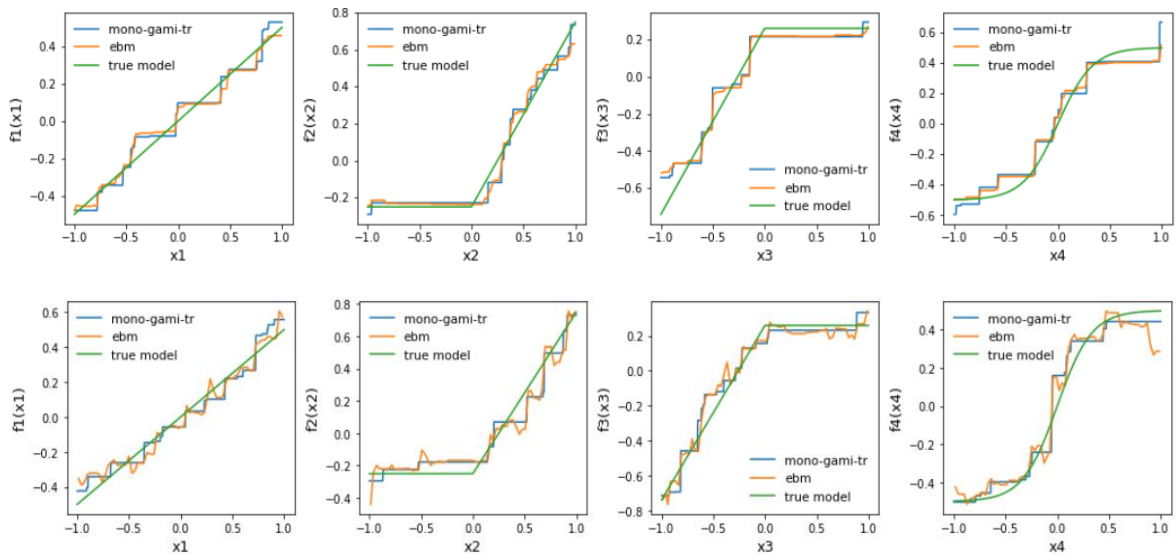

*Figure 1.True and fitted models Continuous (top panels) and Binary (bottom panels) cases*

The model performance on testing set is given in Table 1 (Since model performance is not our main goal, we have just used a single run). The two algorithms have similar performance for all four simulation cases, so the restriction to monotonicity does not come at a price. In this case, the data generation mechanism is based on monotone models. If this is not the case, restrictions to monotone models can cause some loss in performance.

Next, we compare the fitted models. The top panels in Figure 1 show the true and fitted models for continuous responses. The green line is the true model. The fits for both EBM (orange) and mono-GAMI-Tree (blue) are close to each other. There is only slight non-monotonicity in EBM while mono-GAMI-Tree is fully monotone as it should be. The bottom panels show the results for the binary case. Again, both EBM and mono-GAMI-Tree approximate the true models (green) reasonably well, the degree of approximation is not as good as the continuous case. Both exhibit more jumpiness. This is to be expected in binary cases where there is less information. In addition, EBM has more non-monotone behavior compared to the continuous case.

Now we consider the following second order model

$$f(x) = \max(x_1, x_2) + (x_3 + x_4 + x_3 x_4) = g_{12}(x_1, x_2) + g_{34}(x_3, x_4).$$

The model set up and simulation details are the same as those for the first-order model earlier. The model is monotone in each of the four variables (shown in Figures 2-3). Table 2 shows the fitted performance metrics on the test dataset. Again, the performances of EBM and Mono-GAMI-Tree are similar. In addition, it seems Mono-GAMI-Tree has less overfitting compared to EBM, due to the monotonic constraint.

*Table 2. Performance for EBM and monotone GAMI-Tree for Second-Order Model*

| | **Continuous** | | **Binary** | |
|---|---|---|---|---|
| **Algorithm** | **Train (rmse)** | **Test (rmse)** | **Train (AUC)** | **Test (AUC)** |
| mono-GAMI-Tree | 1.987 | 2.013 | 0.741 | 0.725 |
| EBM | 1.973 | 2.015 | 0.751 | 0.725 |

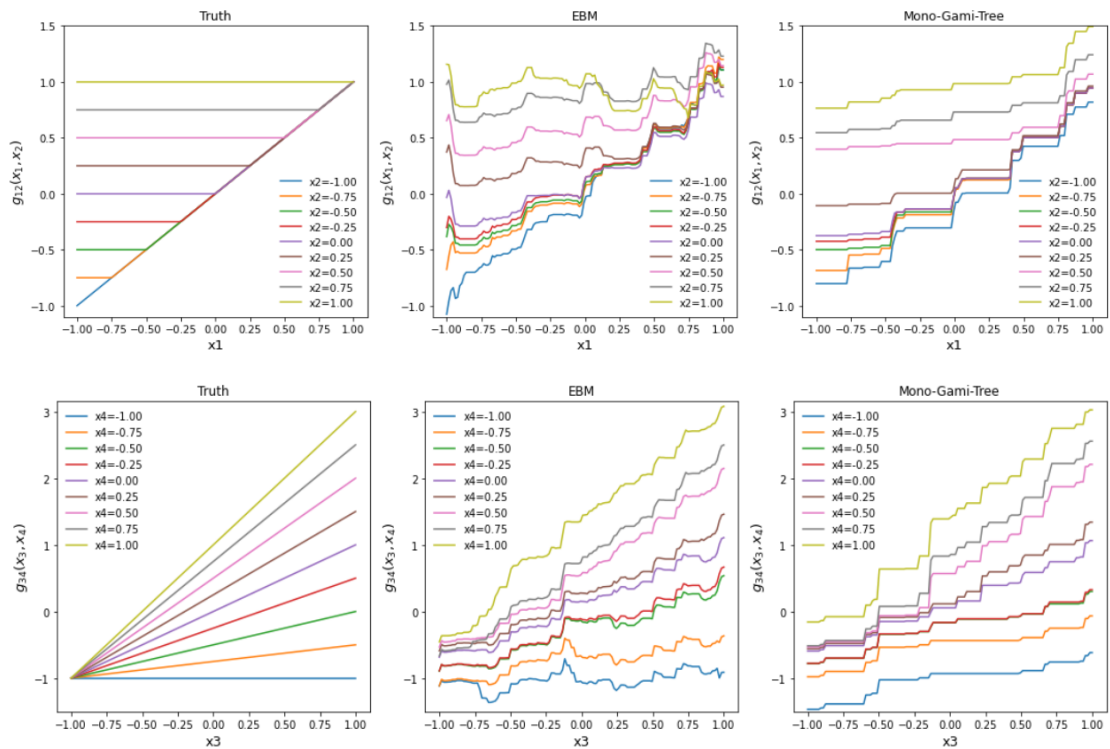

*Figure 2: Continuous response: True (left); fitted: EBM (middle) and mono-GAMI-Tree (right)*
*Top panel $g_{12}(x_1, x_2) = max(x_1, x_2)$; Bottom panel $g_{34}(x_1, x_2) = x_3 + x_4 + x_3 x_4$*

Figure 2 shows fitted model for the continuous response case for both $g_{12}(x_1, x_2) = \max(x_1, x_2)$ and $g_{34}(x_1, x_2) = x_3 + x_4 + x_3 x_4$. While the test RMSEs are both close to the oracle ($\sigma = 2$), the fitted models have some deficiencies for both EBM and Mono-GAMI-Tree. Specifically, the non-monotonic behavior of EBM is more prevalent in this case. While Mono-GAMI-Tree is strictly monotonic, there are jumps in the middle and large spread at the boundary ($x_1 = 1$ and $x_3 = -1$) as opposed to a constant value. Increasing sample size will mitigate these issues.

Figure 3 shows the corresponding results for binary response. The conclusions are similar. We can see the wiggles and non-monotonicity of EBM. For Mono-GAMI-Tree, in addition to the jumpiness and larger spread, the model also seems to be compressed in a tighter range compared to the continuous case.

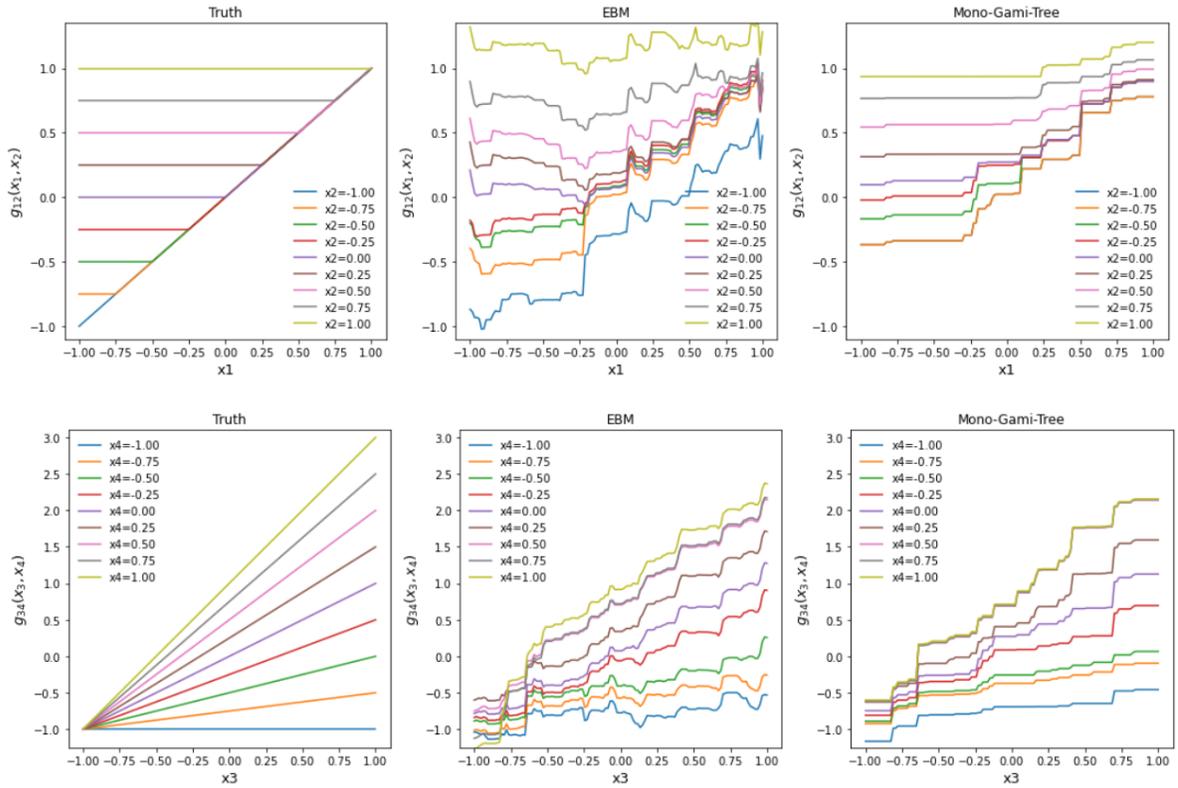

*Figure 3. Binary response: True (left); fitted: EBM (middle) and mono-GAMI-Tree (right). Top panel $g_{12}(x_1, x_2) = \max(x_1, x_2)$; Bottom panel $g_{34}(x_1, x_2) = x_3 + x_4 + x_3 x_4$*

We now consider the behavior of the individual components (main effects and second-order interactions) for the second-order model. EBM uses two-stage modeling approach to separate the main-effects and two-way interactions, whereas monotone GAMI-Tree applies just parsing and purification step. Here we compare the two. Figure 4 presents the pure main effects for the continuous case (top panel) and binary case (bottom panel). The estimated main effects for EBM and monotone GAMI-Tree are similar. They are generally centered around the true main-effect (green line) except at the boundaries. However, the estimated main effects for both models exhibit some non-monotonicity although it is more prevalent for EBM.

Note that the monotonicity requirement is for the full model, and not separately for the main effects or interactions. The pure interaction effect will not be monotonic on either variable; in fact, it will show both increasing and decreasing trend, depending on the value of the other variable (see Figure 5). For main effect, one cannot say in general if it should be monotone or not. To reiterate, **the fitted model is monotonic, but the main-effect component may not be**. In this particular simulation, all variables are independent, so the main effect of $x_j$ is simply $\int \hat{f}(x)dx_{-j}$ which is monotonically increasing **in theory**. However, our numerical approximation to the main effect is not monotonic. This is seen from the minor violations of monotonicity for mono-GAMI-Tree in Figure 4. More generally, if the variables are not independent, one cannot even conclude that the theoretical main effect should be monotonic.

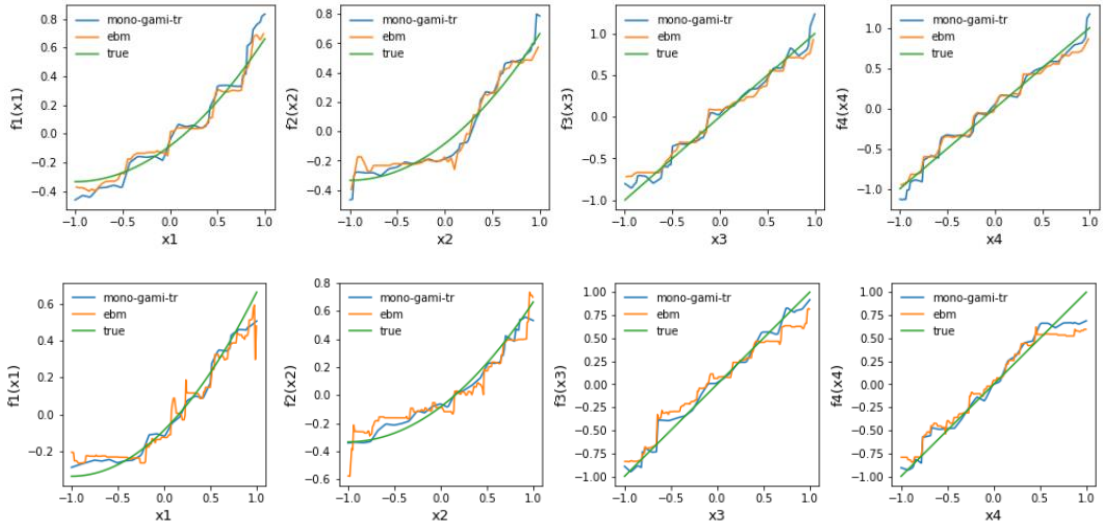

*Figure 4. Fitted pure main effects for Second-Order Model: Continuous case (upper) and Binary case (lower)*

Figure 5 shows the fitted effects of pure interactions from monotone GAMI-Tree and EBM for the continuous case. Except for the behavior near the boundary of $x_j = \pm 1$, the general interaction patterns are similar for both. However, Monotone GAMI-Tree has a tighter range than EBM, whereas EBM has larger cliffs or jumps at the boundary, which indicates overfitting.

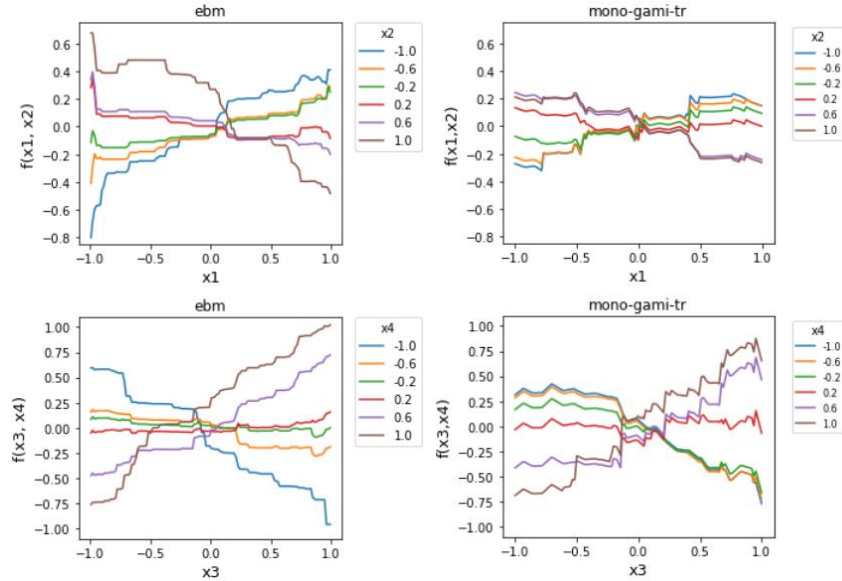

*Figure 5. Fitted pure interaction for Model 2, Continuous case*

## 4. Summary

Interpretability and monotonicity are two very important attributes of model conceptual soundness, especially in highly regulated industries like banking. In this paper, we have proposed a new algorithm which has the same interpretability as EBM and GMI-Lin-T while enjoying monotonicity. Our algorithm is based on xgboost with monotonicity and interaction constraints. We added a parsing and purification step so the "black-box" xgboost model has transparency. We demonstrated the usefulness of this technique through simulation studies. A few questions remain: should the main-effect of a monotonic variable be monotonic? If so, how do we enforce that when performing purification? Nevertheless, our main goal is to guarantee monotonicity and we have achieved this goal. We will investigate the open questions in the future.

# Appendix A: XGBoost Interaction Constraint

Interaction constraint is a standard feature that is supported by Xgboost. However, the documentation on xgboost website gives misleading information on how it works and it is necessary to clarify this confusion here. The documentation says

*For one last example, we use [[0, 1], [1, 3, 4]] and choose feature 0 as split for the root node. At the second layer of the built tree, 1 is the only legitimate split candidate except for 0 itself, since they belong to the same constraint set. Following the grow path of our example tree below, the node at the second layer splits at feature 1. But due to the fact that 1 also belongs to second constraint set [1, 3, 4], at the third layer, we are allowed to include all features as split candidates and still comply with the interaction constraints of its ascendants.*

The documentation suggests that after splitting on feature 0 and 1, the third layer is allowed to split on [1,3,4]. This could result in interaction among 0, 1, 3, if feature 3 is used to split. With similar logic, if we specify two-way interaction constraint [[0, 1], [1, 3]], it will imply that the three-way interaction among 0, 1, 3 may be modeled, which is not what we want. After looking into the source code (xgboost/constraints.cc at master · dmlc/xgboost · GitHub), and doing numerical experiments, we found the documentation is incorrect. The code implementation makes sure the set of splitting variables in any tree branch has to be **fully** contained in one of the constraint sets. Hence there won't be any branch splitting on features 0, 1, and 3. Note different branches within one tree can model different interaction terms. For example, the following tree structure could happen under the constraint [[0, 1], [1, 3]], nevertheless, it still satisfies our two-way interaction constraint because it can be written as

$T(x_0, x_1, x_3) = v_3(x_1 \leq 0)(x_3 \leq 0) + v_4(x_1 \leq 0)(x_3 > 0) + v_5(x_1 > 0)(x_0 \leq 0) + v_6(x_1 > 0)(x_0 > 0)$,

where $v_i$'s are node values at the leaf node $i = 3, 4, 5, 6$.

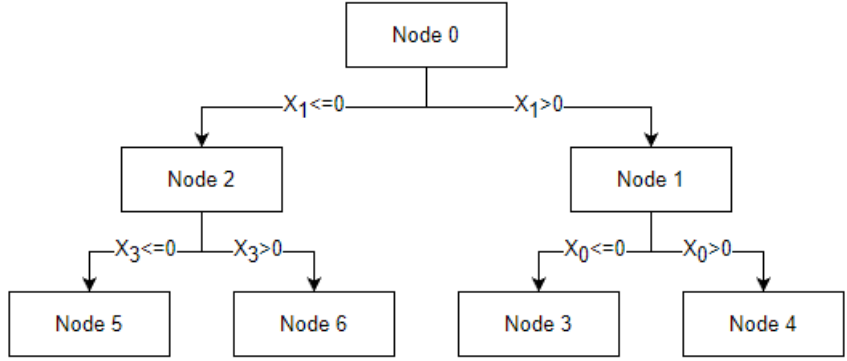

# Appendix B: Residual for binary response

After the main-effects are fitted, residuals need to be calculated to filter top interactions. For continuous response $y = f(x) + \epsilon$, the residuals are simply $y - \hat{f}(x)$ and $E(y - \hat{f}(x)) = f(x) - \hat{f}(x)$ captures the difference between true model and fitted model. For binary response $y \sim Bernoulli(p(x))$, the conventional residual of $y - \hat{p}(x)$ measures the difference in probability space, since $E(y - \hat{p}(x)) = p(x) - \hat{p}(x)$. Since interactions are usually specified in the logit space of $f(x) = \log\frac{p(x)}{1-p(x)}$, a different residual definition measuring $f(x) - \hat{f}(x)$ is needed.

Using first order Taylor expansion, we have $f(x) - \hat{f}(x) = \log\frac{p(x)}{1-p(x)} - \log\frac{\hat{p}(x)}{1-\hat{p}(x)} \approx (p(x) - \hat{p}(x)) \times \frac{1}{\hat{p}(x)(1-\hat{p}(x))} = \frac{p(x)-\hat{p}(x)}{\hat{p}(x)(1-\hat{p}(x))}$. This means we can use $\frac{y-\hat{p}(x)}{\hat{p}(x)(1-\hat{p}(x))}$ as an approximation of $f(x) - \hat{f}(x)$ in the binary case.

Another way of looking at this is provided by xgboost. At each iteration of xgboost, we have the current ensemble model as $\hat{f}(x)$ and the goal is to find tree $T(x)$ such that the loss $\frac{1}{n}\sum_{i=1}^{n} \ell(y_i, \hat{f}(x_i) + T(x_i))$ is minimized. To do this, the loss is expanded through a second-order Taylor expansion: $L \approx \frac{1}{n}\sum_{i=1}^{n} \ell(y_i, \hat{f}(x_i)) + \frac{1}{n}\sum_{i=1}^{n} g_i T(x_i) + \frac{1}{2n}\sum_{i=1}^{n} h_i T(x_i)^2 = C + \frac{1}{2n}\sum_{i=1}^{n} h_i \left(T(x_i) + \frac{g_i}{h_i}\right)^2$, where $g_i = \frac{\partial \ell(y_i, \hat{f}(x_i))}{\partial \hat{f}(x_i)}$, $h_i = \frac{\partial^2 \ell(y_i, \hat{f}(x_i))}{\partial [\hat{f}(x_i)]^2}$ and $C$ is a constant. To minimize loss, it amounts to fit a weighted least square with $-\frac{g_i}{h_i}$ as response and $h_i$ as weights. For binary response, the loss function $\ell(y, f(x)) = yf - \log(1 + e^f)$, so $g_i = y_i - \hat{p}(x_i)$ and $h_i = -\hat{p}(x_i)(1 - \hat{p}(x_i))$, and $-\frac{g_i}{h_i} = \frac{y_i - \hat{p}(x_i)}{\hat{p}(x_i)(1 - \hat{p}(x_i))}$ is the pseudo-residual.